\documentclass[letterpaper]{article}
\usepackage{aaai}
\usepackage{times}
\usepackage{helvet}
\usepackage{courier}

\usepackage{microtype}
\usepackage{times}
\usepackage{soul}
\usepackage{url}
\usepackage[utf8]{inputenc}
\usepackage{graphicx}
\usepackage{amsmath}
\usepackage{amsthm}
\usepackage{booktabs}
\usepackage{algorithm}
\usepackage{algorithmic}
\usepackage{amsfonts}
\usepackage{tabularx}
\usepackage{multirow}
\usepackage{arydshln}
\usepackage{mathtools}
\usepackage{amssymb}
\usepackage{threeparttable}
\usepackage{xcolor}

\DeclareMathOperator*{\argmax}{arg\,max}
\DeclareMathOperator*{\argmin}{arg\,min}

\begin{document}
%
\title{Generating Semantically Valid Adversarial Questions for TableQA}
\author{
    Yi Zhu\textsuperscript{1}\thanks{\enspace Work done as an intern at Amazon.},
    Yiwei Zhou\textsuperscript{2}, Menglin Xia\textsuperscript{2}\\
    
    \textsuperscript{1}Language Technology Lab, University of Cambridge\\
    \textsuperscript{2}Amazon Alexa\\
    
    {\tt yz568@cam.ac.uk}\\ 
    {\tt \{yiweizho,ximengli\}@amazon.co.uk}
}
\maketitle
\begin{abstract}
Adversarial attack on question answering systems over tabular data (TableQA) can help evaluate to what extent TableQA systems can understand natural language questions and reason with tables.
However, generating natural language adversarial questions is difficult, because even a single character swap could lead to huge semantic difference in human perception.
In this paper, we propose SAGE (Semantically valid Adversarial GEnerator), a Wasserstein sequence-to-sequence model for white-box adversarial attack on TableQA systems. 
To preserve meaning of original questions, we apply minimum risk training with S\textsc{imi}L\textsc{e} and entity delexicalization.
We use Gumbel-Softmax to incorporate adversarial loss for end-to-end training. 
Our experiments show that SAGE outperforms existing local attack models on semantic validity and fluency while achieving a good attack success rate.
Finally, we demonstrate that adversarial training with SAGE-augmented data can improve performance and robustness of TableQA systems.
\end{abstract}

\section{Introduction}
Question answering on tabular data (TableQA) is the task of generating answers to natural language questions given tables as external knowledge \cite[\textit{inter alia}]{DBLP:conf/acl/PasupatL15,zhongSeq2SQL2017,DBLP:conf/acml/ChoAHP18}. It has drawn increasing attention in research.
Compared to standard question answering task, TableQA reflects the real world situation better: users interact with a QA system with natural language questions, which requires the QA system to understand the question and reason based on knowledge base in order to provide the correct answer.
State-of-the-art (SOTA) TableQA systems are reported to have exceeded human performance \cite{DBLP:journals/corr/abs-1908-08113,DBLP:journals/corr/abs-1902-01069}.
Despite their impressive results, an important question remains unanswered: Can these systems really understand natural language questions and reason with given tables, or do they merely capture certain statistical patterns in the datasets, which has poor generalizability?

We leverage adversarial examples
to answer this question, as they have been proven to be useful for unbiased evaluation of machine learning systems \cite{DBLP:conf/nips/WerpachowskiGS19} and improving their robustness on various tasks \cite{jia-liang-2017-adversarial,michel-etal-2019-evaluation}.
Adversarial examples were first introduced to computer vision systems by injecting small adversarial perturbations to the input to maximize the chance of misclassification  \cite{DBLP:journals/corr/GoodfellowSS14}.
This is because small perturbations on continuous pixel values does not affect the overall perception of human beings.
Most current works for producing semantic-preserving natural language adversarial examples \cite[\textit{inter alia}]{ebrahimi-etal-2018-hotflip,ren-etal-2019-generating,zhang-etal-2019-generating-fluent} are constrained to local\footnote{We use the term \textit{local} to represent a group of adversarial attacks with word/subword/character level manipulation such as insertion, deletion and substitution.} attack models (\S\ref{sec:bl}) under the assumption that small changes are less likely to lead to large semantic shift from the original sentence.
However, our experiments demonstrate that these changes will impact language fluency or lead to noticeable difference in meaning, which makes the generated adversarial examples invalid.


In this paper, we propose SAGE (Semantically valid Adversarial GEnerator), which generates semantically valid and fluent adversarial questions for TableQA systems.
%
Main contributions of this paper include:
\begin{itemize}
\item SAGE is a novel white-box attack model that can access the gradients of target systems to generate adversarial questions at sequence level for TableQA.
To the best of our knowledge, this work is the first attempt to bridge the gap between white-box adversarial attack and text generation model for TableQA.\footnote{Compared to white-box attack, black-box attack can only query for the model output.}

\item SAGE is based on our proposed stochastic Wasserstein sequence-to-sequence (Seq2seq) model to generate fluent questions.
To improve attack success rate, it incorporates the adversarial loss directly from the target system with Gumbel-Softmax \cite{DBLP:conf/iclr/JangGP17}. 
To tackle the problem of semantic shift, we use delexicalization to tag the entities in the original instances and employ the semantic similarity score S\textsc{imi}L\textsc{e} \cite{wieting-etal-2019-beyond} to guide the model to generate semantically valid questions. 

\item Through our experiments, we demonstrate that our approach can generate more fluent and semantically valid adversarial questions than baselines using local methods (\S\ref{sec:bl}) while keeping high attack success rate (\S\ref{sec:ae},\S\ref{sec:he}). Moreover, SAGE-generated examples can further improve the test performance and robustness of the target QA systems with adversarial training (\S\ref{sec:aug}).
\end{itemize}


\section{Preliminary}
\begin{table}[t]
\begin{threeparttable}
{\footnotesize {\bf Table}:
	\centering
	\begin{tabularx}{\columnwidth}{llXXXX}
	 \toprule
	 \colorbox{lightgray!30}{\textcolor{red}{Rank}} & \colorbox{lightgray!30}{Nation} & \colorbox{lightgray!30}{Gold} & \colorbox{lightgray!30}{Silver} & \colorbox{lightgray!30}{\textcolor{blue}{Bronze}} & \colorbox{lightgray!30}{Total}\\\toprule
	 1 & Russia & 2 & 2 & 2 & 6\\
     2 & France & 1 & 0 & 0 & 1\\
     2 & Hungary & 1 & 0 & 0 & 1\\
     4 & Ukraine & 0 & 1 & 1 & 2\\
     \textcolor{orange}{5} & Bulgaria & 0 & 1 & 0 & 1\\
     6 & Poland & 0 & 0 & \textbf{1} & 1\\\bottomrule
	\end{tabularx}
\textbf{Question}: \colorbox{lightgray!30}{What is the bronze value associated with ranks over 5?}\\
\textbf{SQL query}: \texttt{SELECT} \textcolor{blue}{Bronze} \texttt{WHERE} \textcolor{red}{Rank} \textcolor{green}{$>$} \textcolor{orange}{5}\\
\textbf{Answer}: 1
}
\caption{An example of WikiSQL dataset. Inputs to the TableQA systems are shaded in gray.} \label{tb:wikisql}
\end{threeparttable}
\end{table}

\subsection{Data and Target System}\label{sec:wikisql}
Because of its well-defined syntax and wide usage, using SQL query \cite{zhongSeq2SQL2017} as the answer form for TableQA has advantages over other forms, such as plain text \cite{DBLP:conf/acl/PasupatL15} or operation sequence in table look-up \cite{DBLP:conf/acml/ChoAHP18}. 
In this work, we use the WikiSQL dataset \cite{zhongSeq2SQL2017}, which is one of the largest TableQA datasets consisting of $24,241$ Wikipedia tables and $80,654$ pairs of automatically generated and human-edited question and SQL query (see Table \ref{tb:wikisql} for an example).

TableQA systems on WikiSQL are trained to generate the SQL query and the final answer to a natural language question on a single table using ONLY the table schema (table header) without seeing table content \cite{zhongSeq2SQL2017}.
The final answer is obtained deterministically by executing the generated SQL query on the corresponding relational database. 
SOTA WikiSQL systems \cite{DBLP:journals/corr/abs-1908-08113,DBLP:journals/corr/abs-1902-01069} use pretrained representation models such as BERT \cite{devlin-etal-2019-bert} as encoder, and constrain the output space of SQL query by casting text-to-SQL generation into many classification tasks predicting the slots of SQL keywords and values (see colored slots in Table \ref{tb:wikisql}). 
They have achieved superhuman test performance on WikiSQL with over $80\%$ query accuracy
(Q-Acc = $\frac{\#\text{correct SQL}}{\#\text{test example}}$) and about $90\%$ answer accuracy (A-Acc = $\frac{\#\text{correct answer}}{\#\text{test example}}$).

In this paper, we use SQLova \cite{DBLP:journals/corr/abs-1902-01069} with BERT large encoder (\textbf{SQLova-L}) as our target system,\footnote{It is the only SOTA TableQA system with released code and pretrained model. We use the version without execution-guided decoding at \url{https://github.com/naver/sqlova}.} because its techniques are representative of the SOTA WikiSQL systems.\footnote{At the time of writing, top systems in WikiSQL leaderboard all use BERT-based encoders to perform SQL slot classifications.}
However, it should be noted that our method is target system agnostic and can be applied to any other differentiable WikiSQL system as well.


\subsection{Problem Definition and Local Attack Models} \label{sec:bl}
Given the table schema $\mathbf{t}$ and original question $\mathbf{y}$, a TableQA system is trained to predict the correct slots of the SQL query:
\begin{equation*}
\resizebox{0.99\columnwidth}{!}{
  \begin{minipage}{\columnwidth}
\centering
$
\begin{aligned}
\arg\max_{\mathbf{L} \in \mathbb{L}} p(\mathbf{L}|\mathbf{y}, \mathbf{t}) = \mathbf{L}_{\text{true}}(\mathbf{y})
\end{aligned}
$
\end{minipage}}
\end{equation*}
where $\mathbf{L}$ is the combined set of predicted labels and $\mathbb{L}$ is the set of all possible label combinations.
Ideally, to attack this system, we want to generate an adversarial question $\hat{\mathbf{y}}$ that is semantically valid compared to the original question, but can cause the system to output a wrong answer:
\begin{equation*}
\resizebox{0.99\columnwidth}{!}{
  \begin{minipage}{\columnwidth}
\centering
$
\begin{aligned}
\arg\max_{\mathbf{L} \in \mathbb{L}} p(\mathbf{L}|\hat{\mathbf{y}}, \mathbf{t}) \neq \mathbf{L}_{\text{true}}(\mathbf{y}) \quad
\text{s.t.} \quad \hat{\mathbf{y}} \stackrel{\text{\tiny{semantic valid}}}{=} \mathbf{y}
\end{aligned}
$
\end{minipage}}
\end{equation*}
We define \textit{semantic validity} in the context of WikiSQL as whether $\hat{\mathbf{y}}$ and $\mathbf{y}$ can be expressed by the same SQL query for human.
Maintaining semantic validity is a difficult task, therefore local attack models such as token manipulation are widely used to limit semantic shift.

We apply three white-box local attack models \cite{ebrahimi-etal-2018-hotflip,michel-etal-2019-evaluation} as our local attack baselines, all of which can be formulated as searching the best token embedding $\hat{\mathbf{y}}_i$ in the first order approximation of the adversarial loss $\mathcal{L}_{\text{adv}}$ around the input token embeddings: 
\begin{equation}\label{eq:adv}
\resizebox{0.90\columnwidth}{!}{
  \begin{minipage}{\columnwidth}
\centering
$
\begin{aligned}
&\argmin_{1 \leq i \leq |\mathbf{y}|,\mathbf{\hat{y}}_i\in\mathcal{V}}
[\hat{\mathbf{y}}_i - \mathbf{y}_i]^T\nabla_{\mathbf{y}_i}\mathcal{L}_{\text{adv}}(\hat{\mathbf{y}}, \mathbf{L}_{\text{true}} (\mathbf{y}), \mathbf{t})\\
&\mathcal{L}_{\text{adv}}(\hat{\mathbf{y}}, \mathbf{L}_{\text{true}}(\mathbf{y}), \mathbf{t}) = 
-\!\! \sum_{l \in \mathbf{L}_{\text{true}}(\mathbf{y})}\!\!\log(1 - p(l | \hat{\mathbf{y}}, \mathbf{t}))
\end{aligned}
$
\end{minipage}}
\end{equation}
where $l$ stands for the label in each SQL slot.
Specifically, the three local attack models are: \textbf{Unconstrained}, which searches $\hat{\mathbf{y}}_i$ within the whole embedding space of $\mathcal{V}$; \textbf{kNN}, which constraints the search space within 10 nearest neighbors of the original token embedding; and \textbf{CharSwap}, which swaps or adds a character to the original token to change it to \texttt{<unk>}.

\section{SAGE: Semantically Valid Adversarial Generator}
\begin{figure}[t]
	\centering
	\includegraphics[width=0.99\columnwidth, trim={5.0cm 0 6.0cm 7.5cm}, clip]{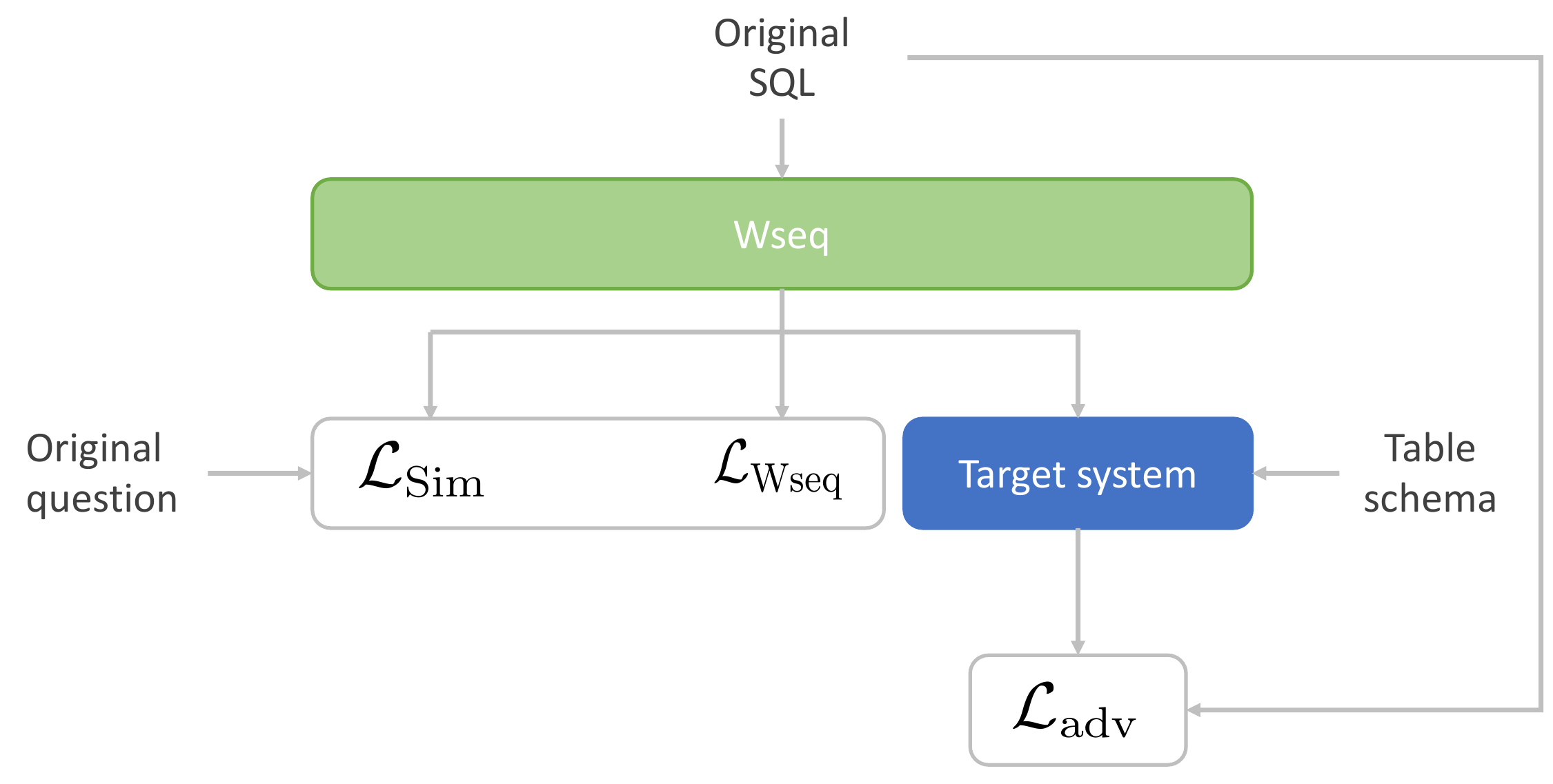}
	\caption{The model architecture of SAGE.}\label{fig:arch}
\end{figure}

Figure \ref{fig:arch} illustrates the model architecture of SAGE with the losses of three main components.
It takes the meaning representation of the question, i.e. SQL query, as input and aims to generate semantically valid and fluent adversarial questions that can fool TableQA systems without changing the gold-standard SQL query. 

We discuss the three main components of SAGE in detail: 1) Stochastic Wasserstein Seq2seq model (Wseq) used for question generation; 2) Delexicalization and minimum risk training with S\textsc{imi}L\textsc{e} (Wseq-S; \cite{wieting-etal-2019-beyond}) to enhance semantic validity; 3) End-to-end training with adversarial loss from the target system using Gumbel-Softmax \cite{DBLP:conf/iclr/JangGP17}.

\subsection{Stochastic Wasserstein Seq2seq Model (Wseq)}\label{sec:wseq}
Wseq is based on Seq2seq model, which encodes the input SQL query $\mathbf{x} = (x_1, x_2, ..., x_{\mid \mathbf{x} \mid})$ with a one layer bidirectional gated recurrent unit (GRU) 
\cite{cho-al-emnlp14},
and the latent representation of the whole SQL sequence, i.e. the last hidden state $\mathbf{z} = [\overrightarrow{\mathbf{h}}_{\mid \mathbf{x} \mid}; \overleftarrow{\mathbf{h}}_1]$ is used to initialize the hidden state of the decoder, which is another one layer GRU.
During each step of decoding, we apply general global attention \cite{D15-1166} and copy mechanism \cite{see-etal-2017-get} to output the predicted token with a Softmax distribution.


Seq2seq model encodes SQL query into a deterministic latent representation $\mathbf{z}$, which can potentially lead to poor generalization during inference.
When an unseen SQL is encoded to a new $\mathbf{z}$, the deterministic Seq2seq model can output nonsensical or unnatural questions from such unseen $\mathbf{z}$, even if it is very close to a training instance \cite{DBLP:conf/conll/BowmanVVDJB16}, which could negatively impact the fluency of the generated questions.
Recent advance in deep generative models based on variational inference \cite{DBLP:journals/corr/KingmaW13} has shown great success in learning smooth latent representations by modeling $\mathbf{z}$ as a distribution to generate more meaningful and natural text \cite{DBLP:conf/conll/BowmanVVDJB16,DBLP:conf/naacl/BahuleyanMZV19}.
To this end, we propose the new Wseq model based on Wasserstein autoencoder \cite{tolstikhin2018wasserstein}, which has been shown to achieve more stable training and better performance than other variational models \cite{DBLP:conf/naacl/BahuleyanMZV19}.
Specifically, Wseq models $\textbf{z}$ as a Gaussian distribution conditioned on the input $\textbf{x}$:
\begin{equation*}
\resizebox{0.99\columnwidth}{!}{
  \begin{minipage}{\columnwidth}
\centering
$
\begin{aligned}
&q(\mathbf{z}|\mathbf{x}) = \mathcal{N}(\boldsymbol{\mu}, \text{diag}\boldsymbol{\sigma}^2)\\
&\boldsymbol{\mu} = \mathbf{W}_{\mu}[\overrightarrow{\mathbf{h}}_{\mid \mathbf{x} \mid}; \overleftarrow{\mathbf{h}}_1] + \mathbf{b}_{\mu}\\
&\log\boldsymbol{\sigma} = \mathbf{W}_{\sigma}[\overrightarrow{\mathbf{h}}_{\mid \mathbf{x} \mid}; \overleftarrow{\mathbf{h}}_1] + \mathbf{b}_{\sigma}
\end{aligned}
$
\end{minipage}}
\end{equation*}
The training objective is to minimize the expected reconstruction loss regularized by the Wasserstein distance $D_{\mathbf{z}}(q(\mathbf{z}), p(\mathbf{z}))$ between the aggregated posterior $q(\mathbf{z})$ and a normal prior $\mathbf{z} \sim \mathcal{N}(\mathbf{0}, \mathbf{I})$:
\begin{equation*}
\resizebox{0.99\columnwidth}{!}{
  \begin{minipage}{\columnwidth}
\centering
$
\begin{aligned}
\mathcal{J} &= -\mathbb{E}_{p(\mathbf{x}, \mathbf{y})q(\mathbf{z}|\mathbf{x}, \mathbf{y})}[\log p(\mathbf{y}|\mathbf{z}, \mathbf{x})] \!+\! D_{\mathbf{z}}(q(\mathbf{z}), p(\mathbf{z}))\\
&\triangleq -\mathbb{E}_{p(\mathbf{x}, \mathbf{y})}\mathbb{E}_{q(\mathbf{z}|\mathbf{x})}[\log p(\mathbf{y}|\mathbf{z}, \mathbf{x})] \!+\! D_{\mathbf{z}}(q(\mathbf{z}), p(\mathbf{z}))
\end{aligned}
$
\end{minipage}}
\end{equation*}
by assuming $\mathbf{y} = \mathbf{y}(\mathbf{x})$, i.e. $\mathbf{y}$ is a function of $\mathbf{x}$ in variational encoder-decoder \cite{DBLP:conf/conll/ZhouN17}.
We use the maximum mean discrepancy (MMD) with the inverse multiquadratic kernel $k(x, y) = \frac{C}{C + \|x - y\|^2_2}$ as $D_{\mathbf{z}}$.
During training, we sample $\mathbf{z}$ from $q(\mathbf{z}|\mathbf{x})$ and approximate MMD with the samples in each mini-batch, so the loss function for each batch can be written as:
\begin{equation}\label{eq:wae}
\resizebox{0.90\columnwidth}{!}{
  \begin{minipage}{\columnwidth}
$
\begin{aligned}
&\!\!\!\!\!\mathcal{L}_{\text{Wseq}} = -\!\!\!\!\sum_{(\mathbf{x}, \mathbf{y}) \in \mathcal{B}}\!\!\sum_{i}^{|\mathbf{y}|} \log p(y_{i}|\mathbf{z}, \mathbf{y}_{< i}, \mathbf{x}) + \lambda_{\text{Wseq}} \hat{D}_{\mathbf{z}}(q(\mathbf{z}), p(\mathbf{z}))\\
&\!\!\!\!\!\hat{D}_{\mathbf{z}}(q(\mathbf{z}), p(\mathbf{z})) = \sum_{i \neq j}\frac{k(\mathbf{z}_i, \mathbf{z}_j) + k(\tilde{\mathbf{z}}_i, \tilde{\mathbf{z}}_j)}{n(n - 1)} - 2\sum_{i, j}\frac{k(\mathbf{z}_i, \tilde{\mathbf{z}_j})}{n^2}\\
\end{aligned}
$
\end{minipage}}
\end{equation}
where $n$ is the size of the batch $\mathcal{B}$, $\tilde{\mathbf{z}}$ is sampled from the posterior $q$, $\mathbf{z}$ is sampled from the prior $p$, and $\lambda_{\text{Wseq}} > 0$ is a hyperparameter controlling the degree of regularization.\footnote{Without causing ambiguity, we use $y_i$ to denote the original token and $\mathbf{y}_i$ for the embedding of this token.}

\begin{table*}[t]
    {
	\centering
	\begin{tabularx}{\textwidth}{l r XXXX XX X}
		&& \textbf{BLEU} & \textbf{METEOR} & \textbf{S\textsc{im}L\textsc{e}} & 
		\textbf{Ecr (\%)} & \textbf{Qfr (\%)} & \textbf{Afr (\%)} & \textbf{Perplexity}\\ \cmidrule(lr){3-6} \cmidrule(lr){7-8} \cmidrule(lr){9-9}
		& Original Questions& - & - & - & - & - & - & 816\\\cmidrule(lr){1-9}
		\multirow{3}{*}{\rotatebox{90}{\footnotesize{Local}}} 
	     
		&Unconstrained      & 79.26 & 51.93 & 87.35 & 100   & 49.46 & 41.23 & 1596\\	
		&kNN                & 80.39 & 56.03 & 93.30 & 100   & 23.80 & 18.23 & 1106\\
		&CharSwap           & 80.76 & 53.91 & 90.51 & 100   & 26.10 & 22.09& 2658\\\cmidrule(lr){1-9}
		
		\multirow{5}{*}{\rotatebox{90}{\footnotesize{Seq2seq-based}}} & Seq2seq w/o delex           & 32.69         & 35.77	        & 80.09         & 68.97         & 12.62 & 11.25 & 515\\
		&Seq2seq                     & 34.91         & 37.58         & 82.79         & 99.38         & 8.98  & 6.69 & 561\\
		&\textbf{Wseq (ours)}                        & 33.72	        & 37.70         & 82.18         & 98.91         & 8.37  & 6.91 & \textbf{474}\\ 
		&\textbf{Wseq-S (ours)}  & \textbf{36.05}& \textbf{37.94}& \textbf{84.32}& \textbf{99.46}& 7.76  & 6.14 & 610\\
		
	    &\textbf{SAGE (ours)}                       & 33.54         & 36.35         & 82.38         & 99.11         & \textbf{17.61} & \textbf{14.46} & 710\\     
		\bottomrule 
	\end{tabularx}}
     \caption{Automatic evaluation metrics on WikiSQL test set for generated adversarial questions. \textbf{Delex} represents entity delexicalization. \textbf{Ecr}, \textbf{Qfr} and \textbf{Afr} represent sequence-level entity coverage rate, query and answer flip rate. The best scores for Seq2seq-based models are in bold.}
     \label{tb:ae}
\end{table*}

\subsection{Enhancing Semantic Validity}\label{sec:ss}
To enhance the semantic validity of the generated questions, we introduce entity delexicalization to improve the sequence-level entity coverage rate and employ S\textsc{imi}L\textsc{e} in minimum risk training to improve the semantic similarity between the generated and the original questions. 

\paragraph{Entity Delexicalization}
If a generated question does not contain all the entities in the WHERE columns of the SQL query, it must be semantically invalid, because all current TableQA systems rely on entity information to locate the correct cells in table to perform reasoning.
To address this problem, we delexicalize the entities appearing in WHERE columns of the SQL query and its corresponding question in WikiSQL.
We replace the $i$-th entity with \texttt{et\_i} in the query/question.
Delexicalization can dramatically reduce the length of the entity tokens our model needs to predict and improve the entity coverage at sequence level.

\paragraph{Minimum Risk Training with S\textsc{imi}L\textsc{e} (Wseq-S)}
How to preserve the original semantics is the main challenge in natural language adversarial attacks.
While including human judgement in the loop is beneficial \cite{jia-liang-2017-adversarial,DBLP:journals/corr/abs-1910-14599},
it is expensive and time consuming. 
Instead, we opt for S\textsc{imi}L\textsc{e} \cite{wieting-etal-2019-beyond}, an automatic semantic similarity score between two sentences to guide our model training, which correlates well with human judgement.
S\textsc{imi}L\textsc{e} calculates the cosine similarity between embeddings of two sentences trained on a large amount of paraphrase text \cite{wieting-gimpel-2018-paranmt}. We choose S\textsc{imi}L\textsc{e} over string matching based metrics, such as BLEU \cite{DBLP:conf/acl/PapineniRWZ02}, because our generated question can be very different in lexical or syntactic realizations from the original question while keeping high semantic similarity.
In order to incorporate S\textsc{imi}L\textsc{e} into our model, we follow \cite{wieting-gimpel-2018-paranmt} and apply minimum risk training \cite{shen-etal-2016-minimum} on a set of generated adversarial questions, i.e. a hypothesis set $\mathcal{H}(\mathbf{x})$, to approximate the whole generated question space given the SQL query x:
\begin{equation}\label{eq:mr}
\resizebox{0.9\columnwidth}{!}{
  \begin{minipage}{\columnwidth}
\centering
$
\begin{aligned}
&\mathcal{L}_{\text{sim}}(\mathbf{x}, \mathbf{y}) = \mathbb{E}_{p(\mathbf{\hat{y}}|\mathbf{x})}[1 - \text{S\textsc{imi}L\textsc{e}}(\mathbf{y}, \mathbf{\hat{y}})]\\
&\triangleq \sum_{\mathbf{\hat{y}} \in \mathcal{H}(\mathbf{x})} (1 - \text{S\textsc{imi}L\textsc{e}}(\mathbf{y}, \mathbf{\hat{y}})) \frac{p(\mathbf{\hat{y}}|\mathbf{x})}{\sum_{\mathbf{\hat{y}}'\in \mathcal{H}(\mathbf{x})}p(\mathbf{\hat{y}}'|\mathbf{x})}
\end{aligned}
$
\end{minipage}}
\end{equation}

\subsection{End-to-end Training with Adversarial Loss}\label{sec:adv}
In order to apply white-box attack, we employ end-to-end training by sending the generated questions to the target system and back-propagate the adversarial loss through our model.
The adversarial loss from the TableQA system, shown in Equation \ref{eq:adv}, maximizes the probability of the target system making incorrect predictions.
However, it is not possible to directly back-propagate the adversarial loss to our attack model through the discrete question tokens which are generated by operating $\argmax$ on Softmax.
To address the issue, we adopt the \textit{Gumbel-Softmax} \cite{DBLP:conf/iclr/JangGP17} to replace Softmax:
\begin{equation*}
\resizebox{0.99\columnwidth}{!}{
  \begin{minipage}{\columnwidth}
\centering
$
\begin{aligned}
p(y_i) = \frac{\exp((\log(\pi_i) + g_i) / \tau)}{\sum_{j}^{|\mathcal{V}|}\exp((\log(\pi_j) + g_j) / \tau)}
\end{aligned}
$
\end{minipage}}
\end{equation*}
where $\pi_i$ is the probability after Softmax for token $i$ in the output vocabulary with size $|\mathcal{V}|$, $g_i$ is the Gumbel($0$, $1$) distribution sample, and $\tau$ controls the smoothness of the distribution.
We still use $\argmax$ to discretize $y_i$ at each time step during generation, but approximate the backward gradients with the Straight-Through (ST) \cite{DBLP:journals/corr/BengioLC13} Gumbel estimator to enable end-to-end training.

Finally, for each batch, SAGE combines the losses of previous three components all together:
\begin{equation*}
\resizebox{0.99\columnwidth}{!}{
  \begin{minipage}{\columnwidth}
\centering
$
\begin{aligned}
\mathcal{L} = \mathcal{L}_{\text{Wseq}} +
&\sum_{(\mathbf{x}, \mathbf{y}, \mathbf{t}) \in \mathcal{B}}
\Big[\lambda_{\text{sim}}\mathcal{L}_{\text{sim}}(\mathbf{x}, \mathbf{y})\\ 
&+ \sum_{\mathbf{\hat{y}} \in \mathcal{H}(\mathbf{x})} \lambda_{\text{adv}}\mathcal{L}_{\text{adv}}\big(\hat{\mathbf{y}}, \mathbf{x}, \mathbf{t}\big)\Big]
\end{aligned}
$
\end{minipage}}
\end{equation*}
where $\mathcal{L}_{\text{Wseq}}$, $\mathcal{L}_{\text{sim}}$ and $\mathcal{L}_{\text{adv}}$ are losses from Equation \ref{eq:wae}, \ref{eq:mr} and \ref{eq:adv}, and $\lambda_{\text{sim}}$ and $\lambda_{\text{adv}}$ are hyperparameters fixed to $0.8$ and $0.1$ respectively according to the development set scores on a spectrum of automatic evaluation metrics to be introduced in \S\ref{sec:ae}.
We use Adam \cite{DBLP:journals/corr/KingmaB14} as the optimizer with the learning rate of $0.001$.
The objective of SAGE takes the text fluency, semantic validity, and the adversarial attack into consideration.

\section{Experiments}\label{sec:exp}

\subsection{Automatic Evaluation} \label{sec:ae}
We evaluate the generated questions in three aspects: semantic validity, flip rate, and fluency.
\paragraph{Semantic Validity} 
As discussed in \S\ref{sec:ss}, the generated adversarial question is semantically valid only if it contains all required entities and preserves the meaning of original question.
We use sequence-level entity coverage rate (\textbf{Ecr}) and semantic similarity to evaluate semantic validity.
Ecr is defined as the ratio between the number of generated questions with all required entities ($v$) and the total number of generated adversarial questions ($m$): $\text{Ecr} = v/m$.
To measure semantic similarity, We use \textbf{BLEU}, \textbf{METEOR} \cite{DBLP:conf/acl/BanerjeeL05} and \textbf{S\textsc{im}L\textsc{e}}.
BLEU is based on exact n-gram matching, so it does not give any credit to semantically similar sentences different in lexical realizations.
METEOR computes the unigram matching F-score using stemming, synonymy and paraphrasing information, allowing for certain lexical variations. 
S\textsc{im}L\textsc{e} is the only embedding-based similarity metric free from string matching.

\paragraph{Flip Rate}
WikiSQL systems are evaluated with both query accuracy and answer accuracy (\S\ref{sec:wikisql}). Correspondingly, we use query flip rate (\textbf{Qfr}) and answer flip rate (\textbf{Afr}) to measure the attack success of our generated questions.
Out of $v$ valid adversarial questions, $l$ questions cause SQL query error and $a$ for answer error, so Qfr and Afr can be calculated as $\text{Qfr} = l/m$ and $\text{Afr} = a/m$.

\paragraph{Fluency}
We want the generated questions to be fluent and natural.
Following \cite{Dathathri2020Plug}, we use GPT-2 based word-level \textbf{perplexity} \cite{radford2019language} as an automatic evaluation metric for fluency.
Fluency is different from semantic validity, because if a generated question differs in meaning from its original question (not semantic valid), it can still be completely fluent to human.

We compare SAGE with two groups of baselines.
The first group comprises the three local attack models discussed in \S\ref{sec:bl}.
For fair comparison with SAGE which employs entity delexicalization, we apply the same technique during local attack so that all the entities in the original question will be preserved.\footnote{Local attack models without entity delexicalization has much worse Ecr, Qfr and Afr, e.g. 24.47 Qfr for Unconstrained compared to 49.46 in Table \ref{tb:ae}.}
The second group includes Seq2seq-based models, i.e., our Wseq and Wseq-S with delexicalization, as well as deterministic Seq2seq models with and without entity delexicalization, both of which have exactly the same model architecture as Wseq, except for the stochastic component.


Table \ref{tb:ae} shows the results of automatic evaluation.
Since local models only make changes to a single token, compared to Seq2seq-based models, they can easily achieve much higher scores in BLEU and METEOR, which are based on string matching. 
However, such gap is reduced significantly between local and Seq2seq-based models in S\textsc{im}L\textsc{e}.
This suggests that although
questions generated by Seq2seq-based models are different in textual realization from the original questions, the semantic meaning is greatly preserved.
Since S\textsc{im}L\textsc{e} is still based on the bag-of-words assumption that ignores syntactic structure, local models inevitably have higher scores than those generated by Seq2seq-based models from scratch, which contain more significant lexical and syntactic variations, examplified in qualitative analysis (\S\ref{sec:qa}).
We argue that these variations are beneficial because they mimic the diversity of human language.
Additionally, despite the high flip rates, local models obtain unreasonably high perplexities, suggesting that many adversarial questions generated by them tend to be non-fluent or nonsensical, which is further confirmed in human evaluation (\S\ref{sec:he}).

Among Seq2seq-based models, Wseq is the most fluent model with the lowest perplexity, demonstrating the effectiveness of our stochastic Wasserstein encoder-decoder.
Wseq-S performs the best in terms of semantic validity whereas Seq2seq w/o delex ranks the worst.
This proves the usefulness of entity delexicalization and minimum risk training with S\textsc{im}L\textsc{e} in capturing the meaning of original question, which is also evidenced by their similar perplexities.
While maintaining comparable semantic validity and fluency,\footnote{SAGE's lower semantic validity compared to Wseq-S suggests that the objectives of adversarial loss and semantic validity are not perfectly aligned.} SAGE achieves the highest flip rate outperforming all Seq2seq baselines, with a 9.85 absolute Qfr increase over Wseq-S.
It shows that SAGE can generate much more adversarial questions with high semantic validity and fluency, which is the main focus of this paper.
Moreover, SAGE-generated adversarial questions are further proven to improve the performance and robustness of the target system (\S\ref{sec:aug}).

\begin{table}[t]
\begin{threeparttable}
	{
	\centering
	\begin{tabularx}{\columnwidth}{r XX}
		& \textbf{Validity (\%)}$\uparrow$ & \textbf{Fluency~(rank)}$\downarrow$ \\\toprule
		Original Questions                   	& - & 2.2\\\cmidrule(lr){1-3}
		
		Unconstrained               & 20.3 & 4.39\\
		kNN                         & 64.0 & 3.39\\\cmidrule(lr){1-3}
		Seq2seq w/o delex           & 78.7 & 2.99\\	
		Seq2seq                     & 89.3 & 2.56\\
		\textbf{Wseq (ours)}                        & 88.7 & \textbf{2.42}\\
	    \textbf{Wseq-S (ours)} & \textbf{90.3} & 2.61\\
	    \textbf{SAGE (ours)}                       & 78.7$^\dagger$ & 2.71$^\ddagger$\\
		\bottomrule       
	\end{tabularx}}
{\footnotesize
$^\dagger$: Significant compared to kNN ($p < 0.01$).\\
$^\ddagger$: Significant compared to kNN ($p < 0.01$) and Seq2seq w/o delex ($p < 0.05$).
}
\caption{Human evaluation results on $100$ questions per task from each model. \textbf{Validity} is the percentage of semantically valid questions. \textbf{Fluency} is the mean rank of each model over sampled questions. The best scores are in bold. The Fleiss' Kappa for inter-annotator agreement is 61.0, which falls in the interval of substantial agreement.} \label{tb:he}  
\end{threeparttable}
\end{table}

\subsection{Human Evaluation} \label{sec:he}
Due to the limitations of automatic metrics in evaluating semantic validity and fluency, we introduce human evaluation to substantiate our findings.
We sample 100 questions from the WikiSQL test set and recruit three native expert annotators to annotate the adversarial examples generated by each model in the tasks of semantic validity and fluency.
Here we focus on the Seq2seq-based models as well as the best-performed local attack models from Table~\ref{tb:ae}.
\paragraph{Semantic Validity} 
It is extremely difficult and error prone to require annotators to write the corresponding SQL query for generated and original questions to measure semantic validity.
Instead, we ask the annotators to make a binary decision on whether the generated and original question have the same query process given table, i.e. whether they use the same columns and rows of the table for the same answer.
According to \textbf{Validity} in Table \ref{tb:he}, our models substantially outperform the local models. For example, 90.3\% of the generated questions of our Wseq-S are semantically valid, fortifying the effectiveness of entity delexicalization and S\textsc{imi}L\textsc{e}. Also, all Seq2seq-based models achieve higher semantic validity than the local models. 
It demonstrates the capability of Seq2seq-based models in generating more semantically equivalent questions. 
Although local models only introduce minor token-level changes, most of them actually greatly alter the meaning of the original questions and make the generated adversarial questions semantically invalid.
Additionally, it suggests that existing automatic metrics are not sufficient to evaluate semantic validity.
On the other hand, albeit the lower semantic validity than Wseq-S and Wseq, SAGE maintains fairly high score, even over $14\%$ higher than the best local model kNN.
Furthermore, given its high flip rate, SAGE still generates many more semantically valid adversarial questions than any other Seq2seq-based model. 
%

\begin{table}[t]
	\centering
	{
	\begin{tabularx}{\columnwidth}{l XX XX}
		& \multicolumn{2}{c}{\textbf{AdvData-B}} & \multicolumn{2}{c}{\textbf{AdvData-L}}\\\toprule
		& Q-Acc & A-Acc & Q-Acc & A-Acc\\\cmidrule(lr){1-5}
		Before Aug.        & 79.0 & 84.5 & 79.0 & 84.5\\ 
		+30k			   & \textbf{79.5} & 85.2 & 79.3&85.0\\
		+56k		   & 79.4 & \textbf{85.5} & \textbf{79.6} & \textbf{85.3}\\
		\bottomrule       
	\end{tabularx}}
     \caption{WikiSQL test accuracy before and after data augmentation on SQLova with BERT base encoder (SQLova-B). \textbf{Q-Acc} and \textbf{A-Acc} denote query and answer accuracy. 
 \textbf{AdvData-B}  and \textbf{AdvData-L} are adversarial examples generated using two different target systems, SQLova-B and the released SQLova with BERT large encoder.}
     \label{tb:aug_ori} 
\end{table}

\paragraph{Fluency}
To compare the fluency of generated questions for each model, we follow the practice of \cite{DBLP:conf/naacl/NovikovaDR18} and ask annotators to rank a set of generated questions including the original one in terms of fluency and naturalness.
We adopt ranking instead of scoring in measuring fluency, because we care more about comparison of the models than absolute scores.
To facilitate annotation, we additionally provide a coarse-grained three level guideline to the annotators.
Similar to its automatic fluency evaluation (perplexity), Table \ref{tb:he}~(\textbf{Fluency}) shows that Seq2seq-based models outperform local models markedly.
Wseq tops all models in human evaluation, and only ranks behind the original question by a small margin, verifying its ability in improving generation fluency.
SAGE yields good fluency, significantly better than Seq2seq w/o delex and all other local models.
The decrease in fluency of SAGE compared to Wseq are caused by two reasons: (1) Adding the adversarial loss harms text quality; (2) S\textsc{imi}L\textsc{e} drives the model to generate questions closer to the original ones, which increases semantic validity but sacrifices fluency, and it is the same case in Wseq-S.\footnote{Given the experience in semantic validity task, we suspect that the same three annotators can easily guess the original question and tend to mark it as the most fluent, though the questions for both tasks are disjoint. We will investigate such bias in future work.}

Overall, Seq2seq-based models are better than local models in human evaluations. 
Together with automatic evaluations, SAGE is shown to reach the best trade-off among adversarial attack (flip rate), semantic validity and fluency.
Compared to all baselines, SAGE is able to generate the most \textit{real} adversarial questions with good text quality.

\subsection{Adversarial Training with SAGE}\label{sec:aug}
We augment training data for TableQA systems with SAGE-generated adversarial examples, then test the performance and robustness of the retrained systems.
Specifically, we first train a SQLova system with BERT base encoder (\textbf{SQLova-B}) on account of efficiency, different from SQLova-L with BERT large encoder used in previous experiments.
We then use SAGE to attack both SQLova-B and SQLova-L on WikiSQL training data, and obtain the same amount of adversarial examples, denoted \textbf{AdvData-B} and \textbf{AdvData-L} respectively. 
We retrain SQLova-B using the original training data augmented with half (\textbf{30k}) and full (\textbf{56k}) adversarial examples from AdvData-B and AdvData-L.
Table \ref{tb:aug_ori} shows that adding SAGE-generated adversarial examples improves the performance of SQLova-B on the original WikiSQL test data.
Although AdvData-L is not targeted for SQLova-B, it can still boost the performance of SQLova-B.
We further attack the two retrained SQLova-B systems augmented with full adversarial examples using different attack models.
Table \ref{tb:aug_adv} demonstrates both AdvData-B and AdvData-L can help SQLova-B to defend various attacks, with all flip rates decreased.
In summary, SAGE-generated adversarial examples can improve the performance and robustness of TableQA systems, regardless which target system is used for generation.

\begin{table}[t]
	\centering
	{
	\setlength{\tabcolsep}{4pt}
	\begin{tabularx}{\columnwidth}{l XXXXXX}
		&\multicolumn{2}{c}{\textbf{Before Aug.}} & \multicolumn{2}{c}{\textbf{AdvData-B}} & \multicolumn{2}{c}{\textbf{AdvData-L}}\\\toprule
	    
        Attack model& Qfr & Afr & Qfr & Afr & Qfr & Afr\\\midrule
        
        Unconstrained     & 53.97 & 46.07 & 53.46 & 45.15 & \textbf{51.01} & \textbf{43.26}\\
		kNN               & 27.36 & 21.85 & \textbf{25.29} & \textbf{19.83} & 25.57 & 20.51\\
		SAGE              & 16.55 & 12.31 & \textbf{10.30} & \textbf{8.09}  & 14.21 & 12.19\\
         
		\bottomrule       
	\end{tabularx}
	}
     \caption{Test flip rates before and after data augmentation for SQLova-B.} 
\label{tb:aug_adv} 
\end{table}

\section{Qualitative Analysis}\label{sec:qa}
\begin{table*}[t]
	\centering
    \def\arraystretch{0.95}
	{\footnotesize
	\begin{tabularx}{\textwidth}{@{}lllX}
	\toprule
		& \textbf{Question} & \textbf{SQL} & \textbf{H}\\\toprule
        		
		\multirow{7}{*}{\rotatebox{90}{{\scriptsize Semantic Validity}}}
		& What is the sum of wins after 1999 ? ({\scriptsize \textbf{Original}})          & \texttt{SELECT} SUM(Wins) \texttt{WHERE} Year $>$ 1999 & -\\ 
		& What is the sum of wins \textbf{downs} 1999 ? ({\scriptsize \textbf{Unconstrained}})      & \checkmark & N\\
	    & What is the sum of wins after 1999 \textbf{is} ({\scriptsize \textbf{kNN}})              & \texttt{SELECT} Wins \texttt{WHERE} Year $>$ 1999         & N\\ 
        & How many wins in the years after 1999 ? ({\scriptsize \textbf{Seq2seq}})        & \checkmark & Y\\ 
        & What is the total wins for the year after 1999 ? ({\scriptsize \textbf{Wseq}})  & \checkmark & Y\\
        & What is the sum of wins in the year later than 1999 ? ({\scriptsize \textbf{Wseq-S}}) & \texttt{SELECT} COUNT(Wins) \texttt{WHERE} YEAR $>$ 1999 & Y\\
        & How many wins have a year later than 1999 ? ({\scriptsize \textbf{SAGE}})       & \texttt{SELECT} COUNT(Wins) \texttt{WHERE} YEAR $>$ 1999 & Y\\\hdashline
        
		\multirow{7}{*}{\rotatebox{90}{{\scriptsize Fluency}}}	
        & What was the date when the opponent was at South Carolina ? ({\scriptsize \textbf{Original}}) & \texttt{SELECT} Date \texttt{WHERE} Opponent $=$ at South Carolina & 3.0\\
        & What was the date when the \textbf{jord} was at South Carolina ? ({\scriptsize \textbf{Unconstrained}}) & \checkmark & 5.3\\
        & What was the date when the opponent was at South Carolina \textbf{,} ({\scriptsize \textbf{kNN}}) & \checkmark & 4.0\\
        & What date was the opponent at South carolina ? ({\scriptsize \textbf{Seq2seq}}) & \checkmark & 1.3\\
        & What is the date of the game against at South Carolina ? ({\scriptsize \textbf{Wseq}}) & \checkmark & 4.7\\
        & What is the date of the opponent at South Carolina ? ({\scriptsize \textbf{Wseq-S}}) & \checkmark & 4.0\\ 
        & On what date was the opponent at South Carolina ? ({\scriptsize \textbf{SAGE}}) & \checkmark & 1.7\\
		\bottomrule       
	\end{tabularx}}
     \caption{Examples of generated questions with human annotations (\textbf{H}). Substituted tokens in local models are in bold.
    $\checkmark$ means the model predicted SQL is the same as the original SQL.
     A generated question is considered semantically valid if at least two annotators select Yes (\textbf{Y}). For fluency, we show the average rank from all three annotators (the lower the better).} \label{tb:qa} 
\end{table*}

We study some generated examples and analyze SAGE qualitatively with the help of model output and human evaluations in Table \ref{tb:qa}.
In the first example, adversarial questions generated by local models are all semantically invalid, showing their limitation in meaning preservation, especially when the local edit happens at content words.
On the contrary, all Seq2seq-based models can generate valid questions, and SAGE is able to generate questions that are both semantically valid and challenging to the target system.

The second example demonstrates the low fluency of local models, due to the occurrence of nonsensical word or inappropriate punctuation.
Questions from Seq2seq-based models are generally more meaningful and fluent.
Wseq and Wseq-S can generate very fluent questions, except for the prepositional phrase entity ``at South Carolina". This is because by default all delexicalized entities from the SQL query are inferred to be proper nouns.
However, SAGE manages to bypass this pitfall by incorporating another syntactic structure, which is what we would like to achieve for more diverse question generation.
Additionally, when comparing Wseq-based models, the word ``opponent", used by the original question, is generated by models with
S\textsc{imi}L\textsc{e} but not plain Wseq. This suggests that the  S\textsc{imi}L\textsc{e} is encouraging the model to generate words coming from the original question, which gives us a hint on the relatively low fluency of Wseq-S in \S\ref{sec:he}.

\section{Related Work}
Adversarial attack in text domain is more difficult than vision because of its discrete nature, so most existing works are local models that manipulate tokens, under the assumption that small changes are less likely to lead to large semantic shift from the original sentence.
They include swap, deletion, insertion, substitution and repetition of characters \cite{DBLP:journals/corr/HosseiniKZP17,DBLP:conf/iclr/BelinkovB18,ebrahimi-etal-2018-hotflip} or words \cite{DBLP:conf/ijcai/0002LSBLS18,wallace-etal-2019-universal,michel-etal-2019-evaluation} with both black-box and white-box (similar to our baselines in \S\ref{sec:bl}) methods.
To further encourage the semantic validity and fluency of the generated examples, different methods have been investigated by using language models \cite{zhang-etal-2019-generating-fluent,DBLP:conf/acl/ChengJM19} and restricting search space within synonyms \cite{ren-etal-2019-generating,DBLP:journals/corr/abs-1910-12196}.
Despite their high attack success rates, local models are still limited to only lexical variations \cite{alzantot-etal-2018-generating,jang2019adversarial}, making them easy to defend \cite{DBLP:conf/acl/PruthiDL19,DBLP:journals/corr/abs-2005-14424}.

On the other hand, sequence-level adversarial attack allows for higher level diversity, but with more difficulty in maintaining the semantic validity of generated examples.
Works in this line involve black-box methods such as adding validity-irrelevant sentences to confound QA systems \cite{jia-liang-2017-adversarial}, or using paraphrasing while controling syntax via templates \cite{ribeiro-etal-2018-semantically} and paraphrase networks \cite{DBLP:conf/naacl/IyyerWGZ18}.
\cite{DBLP:conf/iclr/ZhaoDS18} learnt the perturbations in the latent representation space using generative adversarial networks (GANs; \cite{DBLP:conf/nips/GoodfellowPMXWOCB14}), and use the output of the perturbed representations for black-box attack.
SAGE shares the same paradigm as sequence-level methods, and explicitly considers validity for higher semantic equivalence and adversarial loss for more effective attack.
The only concurrent work similar to SAGE is \cite{DBLP:journals/corr/abs-2003-10388}, where they combined GANs and Gumbel-Softmax to generate adversarial sequences for sentiment analysis, which only requires validity of sentiment polarity rather than stricter semantic validity as in our task.

\section{Conclusion}
We proposed SAGE, the first sequence-level model for white-box adversarial attack on TableQA systems.
It is a Wasserstein Seq2seq model with entity delexicalization and semantic similarity regularization. 
SAGE includes the adversarial loss with Gumbel-Softmax in training to enforce the adversarial attack.
Experiments showed that SAGE is effective in consolidating semantic validity and fluency while maintaining high flip rate of generated adversarial examples.
Moreover, these examples have been demonstrated to improve TableQA systems' performance in question understanding and knowledge reasoning, as well as robustness towards various attacks.
In future work, we will continue to investigate how the generated adversarial examples can further promote evaluation and interpretation for TableQA systems.

\bibliographystyle{aaai}
\bibliography{references}

\begin{thebibliography}{}

\bibitem[\protect\citeauthoryear{Alzantot \bgroup et al\mbox.\egroup
  }{2018}]{alzantot-etal-2018-generating}
Alzantot, M.; Sharma, Y.; Elgohary, A.; Ho, B.-J.; Srivastava, M.; and Chang,
  K.-W.
\newblock 2018.
\newblock Generating natural language adversarial examples.
\newblock In {\em EMNLP}.

\bibitem[\protect\citeauthoryear{Bahuleyan \bgroup et al\mbox.\egroup
  }{2019}]{DBLP:conf/naacl/BahuleyanMZV19}
Bahuleyan, H.; Mou, L.; Zhou, H.; and Vechtomova, O.
\newblock 2019.
\newblock Stochastic wasserstein autoencoder for probabilistic sentence
  generation.
\newblock In {\em NAACL}.

\bibitem[\protect\citeauthoryear{Banerjee and
  Lavie}{2005}]{DBLP:conf/acl/BanerjeeL05}
Banerjee, S., and Lavie, A.
\newblock 2005.
\newblock {METEOR:} an automatic metric for {MT} evaluation with improved
  correlation with human judgments.
\newblock In {\em ACL Workshop on Intrinsic and Extrinsic Evaluation Measures
  for Machine Translation and/or Summarization}.

\bibitem[\protect\citeauthoryear{Belinkov and
  Bisk}{2018}]{DBLP:conf/iclr/BelinkovB18}
Belinkov, Y., and Bisk, Y.
\newblock 2018.
\newblock Synthetic and natural noise both break neural machine translation.
\newblock In {\em ICLR}.

\bibitem[\protect\citeauthoryear{Bengio, L{\'{e}}onard, and
  Courville}{2013}]{DBLP:journals/corr/BengioLC13}
Bengio, Y.; L{\'{e}}onard, N.; and Courville, A.~C.
\newblock 2013.
\newblock Estimating or propagating gradients through stochastic neurons for
  conditional computation.
\newblock {\em CoRR}.

\bibitem[\protect\citeauthoryear{Bowman \bgroup et al\mbox.\egroup
  }{2016}]{DBLP:conf/conll/BowmanVVDJB16}
Bowman, S.~R.; Vilnis, L.; Vinyals, O.; Dai, A.~M.; J{\'{o}}zefowicz, R.; and
  Bengio, S.
\newblock 2016.
\newblock Generating sentences from a continuous space.
\newblock In {\em CoNLL}.

\bibitem[\protect\citeauthoryear{Cheng, Jiang, and
  Macherey}{2019}]{DBLP:conf/acl/ChengJM19}
Cheng, Y.; Jiang, L.; and Macherey, W.
\newblock 2019.
\newblock Robust neural machine translation with doubly adversarial inputs.
\newblock In {\em ACL}.

\bibitem[\protect\citeauthoryear{Cho \bgroup et al\mbox.\egroup
  }{2014}]{cho-al-emnlp14}
Cho, K.; van Merri{\"{e}}nboer, B.; G{\"{u}}l{\c c}ehre, {\c C}.; Bahdanau, D.;
  Bougares, F.; Schwenk, H.; and Bengio, Y.
\newblock 2014.
\newblock Learning phrase representations using rnn encoder--decoder for
  statistical machine translation.
\newblock In {\em EMNLP}.

\bibitem[\protect\citeauthoryear{Cho \bgroup et al\mbox.\egroup
  }{2018}]{DBLP:conf/acml/ChoAHP18}
Cho, M.; Amplayo, R.~K.; Hwang, S.; and Park, J.
\newblock 2018.
\newblock Adversarial {TableQA}: Attention supervision for question answering
  on tables.
\newblock In {\em ACML}.

\bibitem[\protect\citeauthoryear{Dathathri \bgroup et al\mbox.\egroup
  }{2020}]{Dathathri2020Plug}
Dathathri, S.; Madotto, A.; Lan, J.; Hung, J.; Frank, E.; Molino, P.; Yosinski,
  J.; and Liu, R.
\newblock 2020.
\newblock Plug and play language models: A simple approach to controlled text
  generation.
\newblock In {\em ICLR}.

\bibitem[\protect\citeauthoryear{Devlin \bgroup et al\mbox.\egroup
  }{2019}]{devlin-etal-2019-bert}
Devlin, J.; Chang, M.-W.; Lee, K.; and Toutanova, K.
\newblock 2019.
\newblock {BERT}: Pre-training of deep bidirectional transformers for language
  understanding.
\newblock In {\em NAACL}.

\bibitem[\protect\citeauthoryear{Ebrahimi \bgroup et al\mbox.\egroup
  }{2018}]{ebrahimi-etal-2018-hotflip}
Ebrahimi, J.; Rao, A.; Lowd, D.; and Dou, D.
\newblock 2018.
\newblock {H}ot{F}lip: White-box adversarial examples for text classification.
\newblock In {\em ACL}.

\bibitem[\protect\citeauthoryear{Goodfellow \bgroup et al\mbox.\egroup
  }{2014}]{DBLP:conf/nips/GoodfellowPMXWOCB14}
Goodfellow, I.~J.; Pouget{-}Abadie, J.; Mirza, M.; Xu, B.; Warde{-}Farley, D.;
  Ozair, S.; Courville, A.~C.; and Bengio, Y.
\newblock 2014.
\newblock Generative adversarial nets.
\newblock In {\em NIPS}.

\bibitem[\protect\citeauthoryear{Goodfellow, Shlens, and
  Szegedy}{2015}]{DBLP:journals/corr/GoodfellowSS14}
Goodfellow, I.~J.; Shlens, J.; and Szegedy, C.
\newblock 2015.
\newblock Explaining and harnessing adversarial examples.
\newblock In {\em ICLR}.

\bibitem[\protect\citeauthoryear{He \bgroup et al\mbox.\egroup
  }{2019}]{DBLP:journals/corr/abs-1908-08113}
He, P.; Mao, Y.; Chakrabarti, K.; and Chen, W.
\newblock 2019.
\newblock {X-SQL:} reinforce schema representation with context.
\newblock {\em CoRR}.

\bibitem[\protect\citeauthoryear{Hosseini \bgroup et al\mbox.\egroup
  }{2017}]{DBLP:journals/corr/HosseiniKZP17}
Hosseini, H.; Kannan, S.; Zhang, B.; and Poovendran, R.
\newblock 2017.
\newblock Deceiving google's perspective {API} built for detecting toxic
  comments.
\newblock {\em CoRR}.

\bibitem[\protect\citeauthoryear{Hwang \bgroup et al\mbox.\egroup
  }{2019}]{DBLP:journals/corr/abs-1902-01069}
Hwang, W.; Yim, J.; Park, S.; and Seo, M.
\newblock 2019.
\newblock A comprehensive exploration on {WikiSQL} with table-aware word
  contextualization.
\newblock {\em CoRR}.

\bibitem[\protect\citeauthoryear{Iyyer \bgroup et al\mbox.\egroup
  }{2018}]{DBLP:conf/naacl/IyyerWGZ18}
Iyyer, M.; Wieting, J.; Gimpel, K.; and Zettlemoyer, L.
\newblock 2018.
\newblock Adversarial example generation with syntactically controlled
  paraphrase networks.
\newblock In {\em NAACL}.

\bibitem[\protect\citeauthoryear{Jang \bgroup et al\mbox.\egroup
  }{2019}]{jang2019adversarial}
Jang, Y.; Zhao, T.; Hong, S.; and Lee, H.
\newblock 2019.
\newblock Adversarial defense via learning to generate diverse attacks.
\newblock In {\em ICCV}.

\bibitem[\protect\citeauthoryear{Jang, Gu, and
  Poole}{2017}]{DBLP:conf/iclr/JangGP17}
Jang, E.; Gu, S.; and Poole, B.
\newblock 2017.
\newblock Categorical reparameterization with {Gumbel-Softmax}.
\newblock In {\em ICLR}.

\bibitem[\protect\citeauthoryear{Jia and
  Liang}{2017}]{jia-liang-2017-adversarial}
Jia, R., and Liang, P.
\newblock 2017.
\newblock Adversarial examples for evaluating reading comprehension systems.
\newblock In {\em EMNLP}.

\bibitem[\protect\citeauthoryear{Kingma and
  Ba}{2015}]{DBLP:journals/corr/KingmaB14}
Kingma, D.~P., and Ba, J.
\newblock 2015.
\newblock Adam: {A} method for stochastic optimization.
\newblock In {\em ICLR}.

\bibitem[\protect\citeauthoryear{Kingma and
  Welling}{2014}]{DBLP:journals/corr/KingmaW13}
Kingma, D.~P., and Welling, M.
\newblock 2014.
\newblock Auto-encoding variational bayes.
\newblock In {\em ICLR}.

\bibitem[\protect\citeauthoryear{Liang \bgroup et al\mbox.\egroup
  }{2018}]{DBLP:conf/ijcai/0002LSBLS18}
Liang, B.; Li, H.; Su, M.; Bian, P.; Li, X.; and Shi, W.
\newblock 2018.
\newblock Deep text classification can be fooled.
\newblock In Lang, J., ed., {\em IJCAI}.

\bibitem[\protect\citeauthoryear{Luong, Pham, and Manning}{2015}]{D15-1166}
Luong, T.; Pham, H.; and Manning, C.~D.
\newblock 2015.
\newblock Effective approaches to attention-based neural machine translation.
\newblock In {\em EMNLP}.

\bibitem[\protect\citeauthoryear{Michel \bgroup et al\mbox.\egroup
  }{2019}]{michel-etal-2019-evaluation}
Michel, P.; Li, X.; Neubig, G.; and Pino, J.
\newblock 2019.
\newblock On evaluation of adversarial perturbations for sequence-to-sequence
  models.
\newblock In {\em NAACL}.

\bibitem[\protect\citeauthoryear{Nie \bgroup et al\mbox.\egroup
  }{2019}]{DBLP:journals/corr/abs-1910-14599}
Nie, Y.; Williams, A.; Dinan, E.; Bansal, M.; Weston, J.; and Kiela, D.
\newblock 2019.
\newblock Adversarial {NLI:} {A} new benchmark for natural language
  understanding.
\newblock {\em CoRR}.

\bibitem[\protect\citeauthoryear{Novikova, Dusek, and
  Rieser}{2018}]{DBLP:conf/naacl/NovikovaDR18}
Novikova, J.; Dusek, O.; and Rieser, V.
\newblock 2018.
\newblock Rankme: Reliable human ratings for natural language generation.
\newblock In {\em NAACL}.

\bibitem[\protect\citeauthoryear{Papineni \bgroup et al\mbox.\egroup
  }{2002}]{DBLP:conf/acl/PapineniRWZ02}
Papineni, K.; Roukos, S.; Ward, T.; and Zhu, W.
\newblock 2002.
\newblock {BLEU}: a method for automatic evaluation of machine translation.
\newblock In {\em ACL}.

\bibitem[\protect\citeauthoryear{Pasupat and
  Liang}{2015}]{DBLP:conf/acl/PasupatL15}
Pasupat, P., and Liang, P.
\newblock 2015.
\newblock Compositional semantic parsing on semi-structured tables.
\newblock In {\em ACL}.

\bibitem[\protect\citeauthoryear{Pruthi, Dhingra, and
  Lipton}{2019}]{DBLP:conf/acl/PruthiDL19}
Pruthi, D.; Dhingra, B.; and Lipton, Z.~C.
\newblock 2019.
\newblock Combating adversarial misspellings with robust word recognition.
\newblock In {\em ACL}.

\bibitem[\protect\citeauthoryear{Radford \bgroup et al\mbox.\egroup
  }{2019}]{radford2019language}
Radford, A.; Wu, J.; Child, R.; Luan, D.; Amodei, D.; and Sutskever, I.
\newblock 2019.
\newblock Language models are unsupervised multitask learners.

\bibitem[\protect\citeauthoryear{Ren \bgroup et al\mbox.\egroup
  }{2019}]{ren-etal-2019-generating}
Ren, S.; Deng, Y.; He, K.; and Che, W.
\newblock 2019.
\newblock Generating natural language adversarial examples through probability
  weighted word saliency.
\newblock In {\em ACL}.

\bibitem[\protect\citeauthoryear{Ren \bgroup et al\mbox.\egroup
  }{2020}]{DBLP:journals/corr/abs-2003-10388}
Ren, Y.; Lin, J.; Tang, S.; Zhou, J.; Yang, S.; Qi, Y.; and Ren, X.
\newblock 2020.
\newblock Generating natural language adversarial examples on a large scale
  with generative models.
\newblock {\em CoRR}.

\bibitem[\protect\citeauthoryear{Ribeiro, Singh, and
  Guestrin}{2018}]{ribeiro-etal-2018-semantically}
Ribeiro, M.~T.; Singh, S.; and Guestrin, C.
\newblock 2018.
\newblock Semantically equivalent adversarial rules for debugging {NLP} models.
\newblock In {\em ACL}.

\bibitem[\protect\citeauthoryear{See, Liu, and
  Manning}{2017}]{see-etal-2017-get}
See, A.; Liu, P.~J.; and Manning, C.~D.
\newblock 2017.
\newblock Get to the point: Summarization with pointer-generator networks.
\newblock In {\em ACL}.

\bibitem[\protect\citeauthoryear{Shen \bgroup et al\mbox.\egroup
  }{2016}]{shen-etal-2016-minimum}
Shen, S.; Cheng, Y.; He, Z.; He, W.; Wu, H.; Sun, M.; and Liu, Y.
\newblock 2016.
\newblock Minimum risk training for neural machine translation.
\newblock In {\em ACL}.

\bibitem[\protect\citeauthoryear{Tolstikhin \bgroup et al\mbox.\egroup
  }{2018}]{tolstikhin2018wasserstein}
Tolstikhin, I.; Bousquet, O.; Gelly, S.; and Schoelkopf, B.
\newblock 2018.
\newblock Wasserstein auto-encoders.
\newblock In {\em ICLR}.

\bibitem[\protect\citeauthoryear{Wallace \bgroup et al\mbox.\egroup
  }{2019}]{wallace-etal-2019-universal}
Wallace, E.; Feng, S.; Kandpal, N.; Gardner, M.; and Singh, S.
\newblock 2019.
\newblock Universal adversarial triggers for attacking and analyzing {NLP}.
\newblock In {\em EMNLP-IJCNLP}.

\bibitem[\protect\citeauthoryear{Werpachowski, Gy{\"{o}}rgy, and
  Szepesv{\'{a}}ri}{2019}]{DBLP:conf/nips/WerpachowskiGS19}
Werpachowski, R.; Gy{\"{o}}rgy, A.; and Szepesv{\'{a}}ri, C.
\newblock 2019.
\newblock Detecting overfitting via adversarial examples.
\newblock In Wallach, H.~M.; Larochelle, H.; Beygelzimer, A.;
  d'Alch{\'{e}}{-}Buc, F.; Fox, E.~B.; and Garnett, R., eds., {\em NeurIPS}.

\bibitem[\protect\citeauthoryear{Wieting and
  Gimpel}{2018}]{wieting-gimpel-2018-paranmt}
Wieting, J., and Gimpel, K.
\newblock 2018.
\newblock {P}ara{NMT}-50{M}: Pushing the limits of paraphrastic sentence
  embeddings with millions of machine translations.
\newblock In {\em ACL}.

\bibitem[\protect\citeauthoryear{Wieting \bgroup et al\mbox.\egroup
  }{2019}]{wieting-etal-2019-beyond}
Wieting, J.; Berg-Kirkpatrick, T.; Gimpel, K.; and Neubig, G.
\newblock 2019.
\newblock Beyond {BLEU}:training neural machine translation with semantic
  similarity.
\newblock In {\em ACL}.

\bibitem[\protect\citeauthoryear{Ye, Gong, and
  Liu}{2020}]{DBLP:journals/corr/abs-2005-14424}
Ye, M.; Gong, C.; and Liu, Q.
\newblock 2020.
\newblock {SAFER:} {A} structure-free approach for certified robustness to
  adversarial word substitutions.
\newblock {\em CoRR}.

\bibitem[\protect\citeauthoryear{Zang \bgroup et al\mbox.\egroup
  }{2019}]{DBLP:journals/corr/abs-1910-12196}
Zang, Y.; Yang, C.; Qi, F.; Liu, Z.; Zhang, M.; Liu, Q.; and Sun, M.
\newblock 2019.
\newblock Open the boxes of words: Incorporating sememes into textual
  adversarial attack.
\newblock {\em CoRR} abs/1910.12196.

\bibitem[\protect\citeauthoryear{Zhang \bgroup et al\mbox.\egroup
  }{2019}]{zhang-etal-2019-generating-fluent}
Zhang, H.; Zhou, H.; Miao, N.; and Li, L.
\newblock 2019.
\newblock Generating fluent adversarial examples for natural languages.
\newblock In {\em ACL}.

\bibitem[\protect\citeauthoryear{Zhao, Dua, and
  Singh}{2018}]{DBLP:conf/iclr/ZhaoDS18}
Zhao, Z.; Dua, D.; and Singh, S.
\newblock 2018.
\newblock Generating natural adversarial examples.
\newblock In {\em ICLR}.

\bibitem[\protect\citeauthoryear{Zhong, Xiong, and
  Socher}{2017}]{zhongSeq2SQL2017}
Zhong, V.; Xiong, C.; and Socher, R.
\newblock 2017.
\newblock {Seq2SQL}: Generating structured queries from natural language using
  reinforcement learning.
\newblock {\em CoRR}.

\bibitem[\protect\citeauthoryear{Zhou and
  Neubig}{2017}]{DBLP:conf/conll/ZhouN17}
Zhou, C., and Neubig, G.
\newblock 2017.
\newblock Morphological inflection generation with multi-space variational
  encoder-decoders.
\newblock In {\em CoNLL–SIGMORPHON 2018 Shared Task: Universal Morphological
  Reinflection}.

\end{thebibliography}

\end{document}